\documentclass[10pt,twocolumn,letterpaper]{article}

\usepackage{cvpr}
\usepackage{times}
\usepackage{epsfig}
\usepackage{amsmath}
\usepackage{amssymb}

\usepackage{makecell}
\usepackage{multirow}
\usepackage{subcaption}
\usepackage{floatrow}
\usepackage{float}
\usepackage{bm}

\newfloatcommand{capbtabbox}{table}[][\FBwidth]

\usepackage[pagebackref=true,breaklinks=true,letterpaper=true,colorlinks,bookmarks=false]{hyperref}

\def\cvprPaperID{*****} 
\def\confName{CVPR WAD}
\def\confYear{2023}

\begin{document}

\title{HazardNet: Road Debris Detection by Augmentation of Synthetic Models}

\author{Tae Eun Choe \\
{\tt\small tchoe@nvidia.com}
\and 
Jane Wu\\
{\tt\small janehwu@stanford.edu}
\and
Xiaolin Lin\\
{\tt\small xiaolinl@nvidia.com}
\and
Karen Kwon\\
{\tt\small kkwon@nvidia.com}
\and
Minwoo Park\\
{\tt\small minwoop@nvidia.com}
}

\maketitle

\begin{abstract}
We present an algorithm to detect unseen road debris using a small set of synthetic models. Early detection of road debris is critical for safe autonomous or assisted driving, yet the development of a robust road debris detection model has not been widely discussed. There are two main challenges to building a road debris detector: first, data collection of road debris is challenging since hazardous objects on the road are rare to encounter in real driving scenarios; second, the variability of road debris is broad, ranging from a very small brick to a large fallen tree. To overcome these challenges, we propose a novel approach to few-shot learning of road debris that uses semantic augmentation and domain randomization to augment real road images with synthetic models. We constrain the problem domain to uncommon objects on the road and allow the deep neural network, HazardNet, to learn the semantic meaning of road debris to eventually detect unseen road debris. Our results demonstrate that HazardNet is able to accurately detect real road debris when only trained on synthetic objects in augmented images.
\end{abstract}

\section{Introduction}

\begin{figure}[ht]
    \centering
    \includegraphics[width=\linewidth]{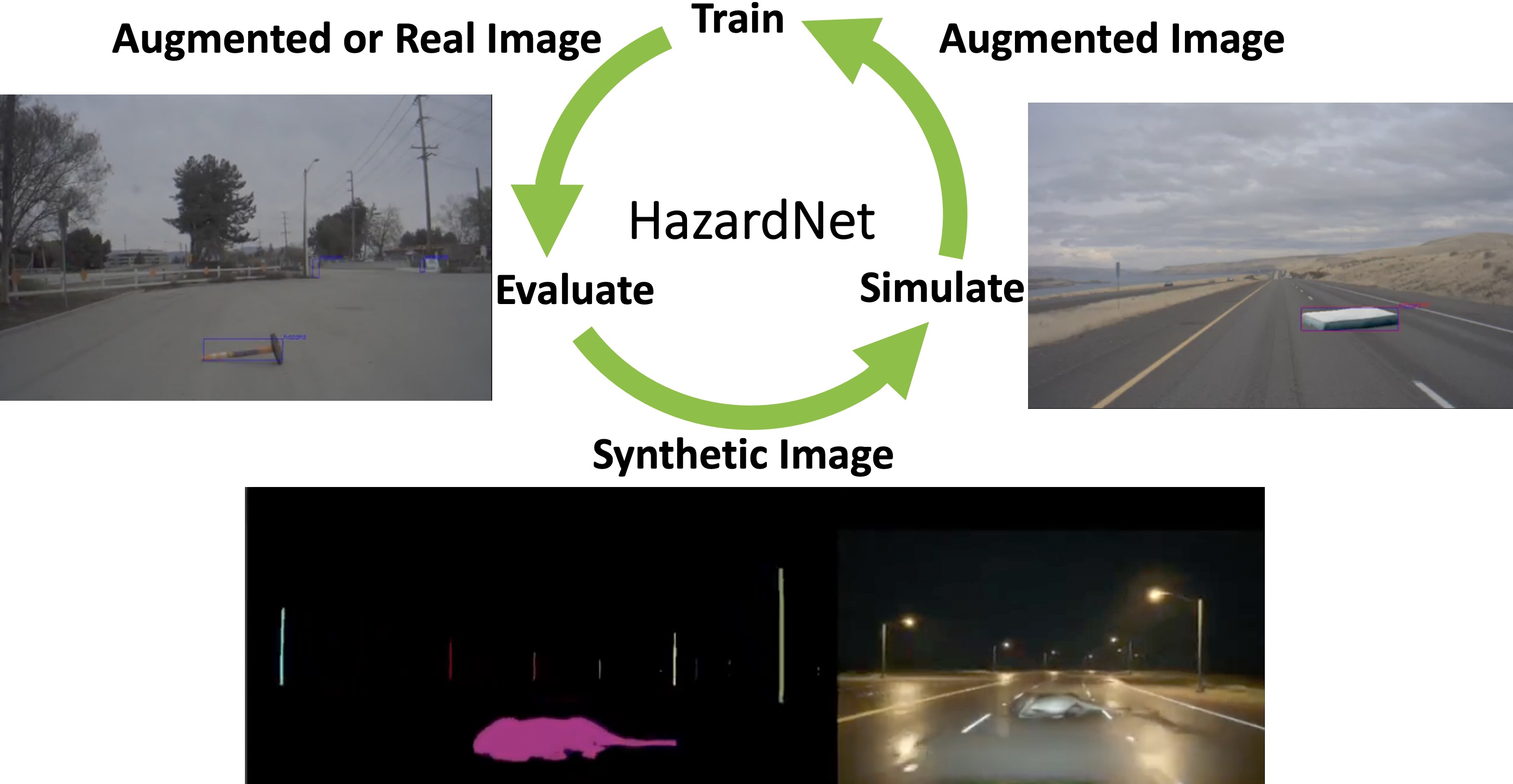}
    \break
    \caption{The cyclic workflow of HazardNet. Synthetic models of road debris are generated with domain randomization and rendered over real images using semantic augmentation. These augmented images are used to train HazardNet and evaluated on augmented and real images of road debris to assess few-shot learning performance. The evaluation results are used to tune the augmented data, and the cycle continues until performance converges.}
    \label{fig:cyclic_flow}
    \vspace{-0.8em}
\end{figure}

Object detection for autonomous vehicles mostly focuses on common objects and obstacles such as vehicles, bikes, pedestrians, traffic light/signs, and lane lines. However, equally important yet challenging objects to detect on the road are hazardous debris such as furniture, tires, fallen trees, animals, potholes, and more. When the height of road debris is larger than 15 centimeters, the treading vehicle may roll or detour from the path and potentially cause a fatal accident. Identification and localization of such hazardous objects is a critical task for both autonomous vehicles and regular road users. Based on the AAA traffic safety report\footnote{\url{https://exchange.aaa.com/prevent-road-debris}}, between 2011-2014, there were 200,000 crashes related to road debris, resulting in 39,000 injuries and 500 deaths. Two thirds of road debris are vehicle parts, unsecured cargo, and separated tow trailers.
Since the most serious accidents occur on highways, where vehicles are traveling at high speeds, it is crucial for road debris detectors to identify objects early from a far distance.

There are two main challenges to road debris detection. The first is that data collection of road debris is extremely time-consuming, expensive, and dangerous. The conventional method requires physical vehicles to collect data. However, road debris are quite rare to encounter and thus requires long data collection times. In addition, when the data collecting vehicle approaches the road debris, it should either stop or detour, which are both dangerous maneuvers on the road. Though staging road debris on private roads is feasible, staging on various public roads is infeasible and limited by regulation. The second challenge is the broad variety of road debris, ranging from a small brick to a large detached trailer. The appearance and shapes of different hazardous road objects are difficult to categorize. For instance, the appearance of a deceased animal and that of a ladder are quite dissimilar. An effective road debris detector should thus be able to identify an enormous number of distinct road objects, making it almost the same as a detector of all objects excluding common road entities. 

We overcome the challenges of limited data and variability of road debris by using semantic augmentation with synthetic models and domain randomization. Contrary to conventional supervised learning methods on real labeled images, we propose to train a machine learning model, HazardNet, to detect hazardous road objects using few-shot learning. We generate training data for this objective by augmenting real images with several representative synthetic models in a semantically valid way. The main goal here is not only detecting road debris in training data but also learning the semantic context of road debris so that instances unseen during training can be detected at test time. This approach avoids dangerous and time-consuming data collection using a physical vehicle and speeds up the development process by using the synthetic data and recycling existing real data as well. See Figure \ref{fig:cyclic_flow}.	 

To learn the general concept of road debris, we apply domain randomization by selecting several representative synthetic objects and placing them on the road in various locations and orientations and in images with different times of day and weather conditions.
When a synthetic model is augmented on the image, it is  augmented not in a random location but on the road as a  hazardous obstacle using ground plane estimation. Since our main focus is to detect debris on the road, the synthetic model is placed on the path where the ego-vehicle moves forward. Our placement method also avoids overlap with common road elements such as vehicles, pedestrians, traffic signs, and vertical poles.  
Both domain randomization and semantic augmentation are essential for few-shot learning of visual and semantic road debris information. Even though there are an uncountable number of distinct road debris, the concept and meaning of road debris can be decoded using these two techniques. 

In this paper, we mostly focus on road debris detection in the autonomous driving domain. However, the approach can be applied to other domains such as general detection of under-represented objects, medical imagery, speech recognition or any machine learning domain utilizing simulation data. 

\begin{figure*}[!htb]
    \centering
    \includegraphics[width=.95\linewidth]{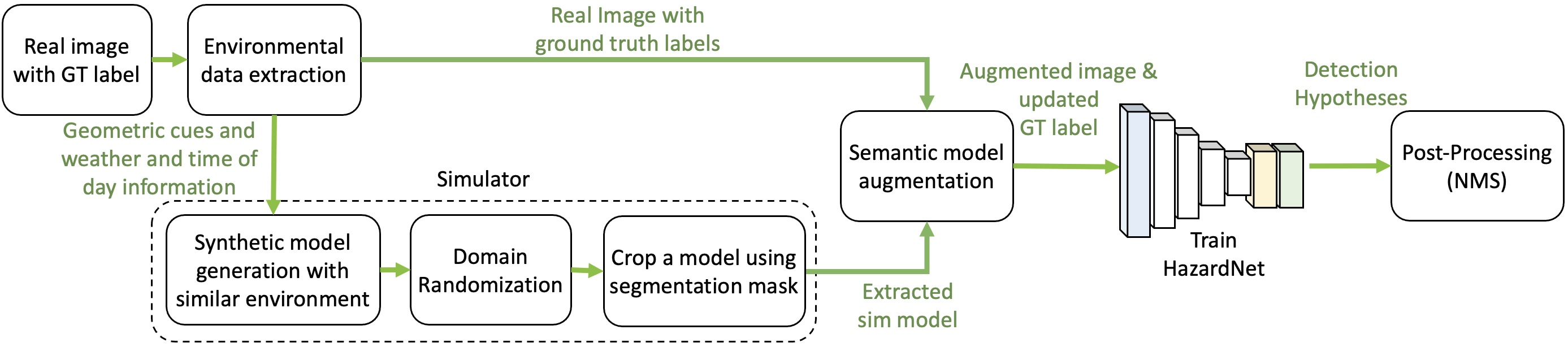}
    \caption{The data generation and training pipeline for HazardNet. Given a real image with ground truth labels, we use geometric cues and environmental information to randomly generate a synthetic model with texture corresponding to a similar environment. Once the model is selected, domain randomization is applied to render the synthetic model with random 3D pose, color tone, and visibility. Next, the rendered model is segmented and used to semantically augment the real image (e.g.\ taking into consideration overlapping objects and lane information.). We use the augmented images to train HazardNet and apply post-processing to the output predictions.}
    \label{fig:workflow}
\end{figure*}

\section{Previous Work}

\textbf{Synthetic Data:} Recent work in deep learning has demonstrated the effectiveness of using synthetic data for object detection \cite{gaidon2016virtual}, semantic segmentation \cite{richter2016playing,ros2016synthia}, lane line detection \cite{GuoCZZMWC20}, and optical flow \cite{ilg2017flownet,mayer2016large}. In the realm of autonomous vehicle perception, fully synthetic datasets such as CARLA \cite{dosovitskiy2017carla}, GTA5 \cite{richter2016playing}, SYNTHIA \cite{ros2016synthia}, and Virtual KITTI \cite{gaidon2016virtual} have been used to improve semantic segmentation of urban scenes \cite{chen2019learning,chen2018road,hoffman2016fcns,lee2018spigan,li2019bidirectional,prakash2019structured,saleh2018effective,tsirikoglou2017procedural,vu2019advent,yue2019domain,zhang2018fully,zhang2020transferring}. A number of these works use adversarial learning to facilitate domain adaptation from simulation to real data \cite{chen2019learning,hoffman2016fcns,lee2018spigan,li2019bidirectional,vu2019advent,zhang2018fully,zhang2020transferring}. More similar to our approach, \cite{tremblay2018training} performs domain randomization by rendering synthetic objects with random background images captured in the real world. However, the generated scenes do not follow any physical constraints, e.g. objects are placed at random 3D locations.

\textbf{Augmentation:}
Rather than using completely synthetic data, several works augment real images by adding synthetic 3D models to the scene. In such cases, augmentation is typically applied to indoor scenes by placing objects in random locations \cite{peng2015learning,su2015render}, and thus the resulting images need not bear resemblance to the physical world.
Though road debris location has a strong association with the image ground plane, existing methods have no mechanism for realistically placing synthetic objects.
More recently, \cite{hinterstoisser2018pre} accomplishes image augmentation by generating a large dataset of 3D CAD models for various household objects and rendering each object over a cluttered background with added Gaussian noise.
Another approach to augmentation is to add masked objects from real images to other images \cite{dwibedi2017cut,georgakis2017synthesizing,rad2017bb8, lee2019copy}.

\textbf{Domain Randomization and Adaptation:}
Domain randomization is one of the most effective methods to reduce the sim-to-real domain gap.
The concept was first introduced in \cite{tobin2017domain} to expose deep neural networks (DNNs) to a wide range of different environments by randomizing the simulator when generating training data.
This approach makes the assumption that if networks are trained on sufficiently varied synthetic data, models trained only in simulation can generalize to the real world without retraining.
As in \cite{tobin2017domain}, domain randomization has most commonly been applied to robot manipulation and control \cite{andrychowicz2020learning,james2017transferring,peng2018sim,sadeghi2016cad2rl,zhang2019adversarial}.
Domain randomization for object detection is another application that has gained interest in the last few years \cite{khirodkar2019domain,prakash2019structured,tremblay2018training,yue2019domain}.
In a similar vein, the objective of domain adaptation is to align the source and target domains such that a model trained on the source domain can be applied to tasks in the target domain \cite{ben2010theory,wang2018deep}. Specifically for object detection, a number of recent works build upon Domain Adaptive Faster R-CNN \cite{chen2018domain}, which adds domain adaptive components to Faster R-CNN \cite{ren2015faster} and trains the model in an adversarial manner.
Subsequently, \cite{he2019multi,hsu2020progressive,raj2015subspace,saito2019strong,zhu2019adapting} have also applied adversarial learning to domain adaptation for object detection.
\cite{xu2020exploring} presents a categorical regularization framework on top of \cite{chen2018domain} that highlights the important image regions corresponding to categorical information.

\textbf{Few-shot Learning:} 
Few-shot, One-shot, or Zero-shot learning extrapolates beyond labeled data by inferring information about instances not seen during training \cite{larochelle2008zero,romera2015embarrassingly,xian2017zero,yu2010attribute}. Recent work in zero-shot object detection \cite{bansal2018zero,demirel2018zero,li2019zero,rahman2018zero,rezaei2020zero,zhu2019zero,zhu2020don} focuses on detecting a set of unseen classes chosen to be excluded during training.
In such cases, high-performance object detection networks, particularly YOLOv2 \cite{redmon2017yolo9000}, have been used as backbones to provide a baseline for detection \cite{demirel2018zero,zhu2019zero,zhu2020don}.
Most recently, \cite{zhu2020don} trained a conditional variational autoencoder to synthesize visual features for an input image, which was then used to re-train a confidence predictor to encourage detection of unseen objects. 

\textbf{Hazard Detection:} Early approaches for image-based road hazard/obstacle detection use classical stereo reconstruction \cite{williamson1998detection,nedevschi2004high} to detect small road obstacles at long distances. Following this work, geometric detection algorithms were used to detect obstacles from stereo images, including extensions of the Stixel algorithm \cite{pfeiffer2011towards}, geometric clustering methods \cite{manduchi2005obstacle,broggi2011stereo}, and the Fast Direct Planar Hypothesis Testing (FPHT) method \cite{pinggera2016lost,ramos2017detecting}, which assumes that the road is planar. However, it is difficult to detect distant shallow objects with stereo vision. In addition, stereo vision and parallax-based approaches on the road are prone to fail and produce numerous false positive detections since road features are ambiguous and the 3D road profile breaks the planar assumption.  More recently, advances in deep learning have shifted the focus to monocular obstacle detection \cite{gupta2018mergenet,ohgushi2020road,sun2020real}. For instance, \cite{ohgushi2020road} proposed an autoencoder framework that leverages semantic segmentation and anomaly calculation to detect obstacle regions in an input image. RGB-D images have also been used as input to DNN-based obstacle detection algorithms, including MergeNet \cite{gupta2018mergenet} and RFNet \cite{sun2020real}. \cite{Di_Biase_2021_CVPR} detects hazardous objects by segmentation -based anomaly detection.  

\section{Few-shot Learning of Road Debris}\label{sec:zero_shot}
The lack of available data naturally makes few-shot learning an attractive approach to detecting real-world road debris.
While we use synthetic road debris to train a DNN and detect those specific models, our ultimate objective is to detect unseen road debris in real-world images. To this end, we apply 1) semantic augmentation to meaningfully place synthetic objects in the scene and 2) domain randomization to increase the variety of road debris appearance. Semantic augmentation and domain randomization endow the DNN with semantic and visual understanding of road debris in the context of the ego vehicle and environmental conditions. 
The data generation and training pipeline is shown in Figure~\ref{fig:workflow}.
Section \ref{sec:3d_model_generation} describes how 3D synthetic models are generated and saved as masked images given an input real-world image. Section \ref{sec:domain_randomization} discusses domain randomization in the context of varying the appearance of the rendered synthetic models. Section \ref{sec:hazardnet} outlines the HazardNet architecture, and Section \ref{sec:metrics} considers important performance metrics for road debris detection.

\subsection{Synthetic model generation}\label{sec:3d_model_generation}
The most common and hazardous road debris are vehicle parts (tires, mufflers, hubcaps, bumpers), unsecured cargo (mattress, furniture, box, detached trailer), tree branches, and animals.
Therefore, we collected 20 synthetic models of the most common road debris, including: cardboard boxes, small and big rocks, tires, wheels, wooden pallets, roadkill, wooden logs, traffic cones, barrels, mattresses, detached mufflers, trash cans, traffic sign bases, and detached trailers. When spawning objects on the road, their 3D information such as location in latitude/longitude, orientation (yaw/pitch/roll), lighting, weather condition, and time of day are saved as metadata. Since the appearance of the synthetic model should match the environmental conditions of the real image being augmented, we ensure that the weather conditions and times of day are consistent. The object is segmented using the instance segmentation mask generated by the simulator and saved as an image along with the mask and metadata.

\begin{figure}[ht]
    \centering
    \includegraphics[width=\linewidth]{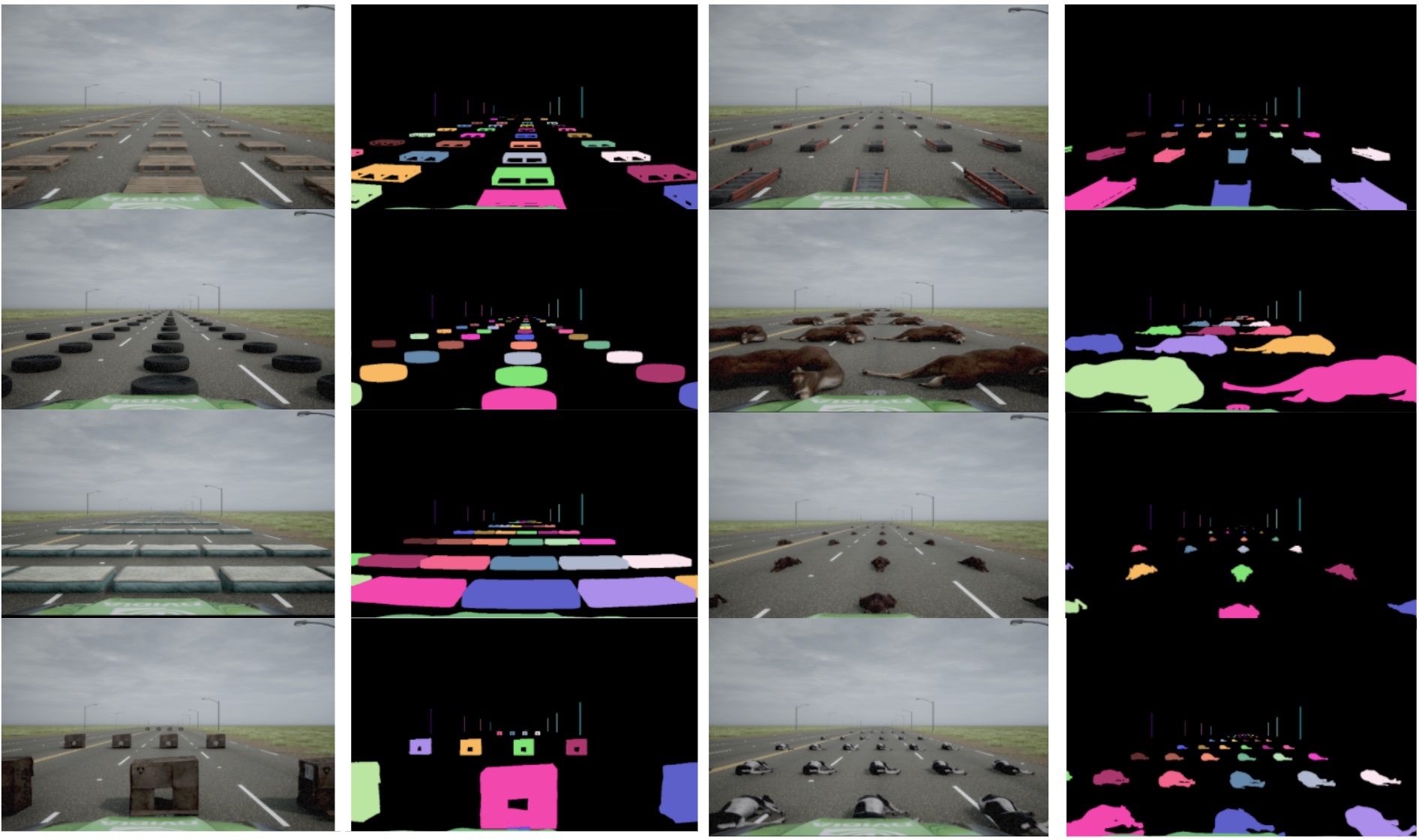}
    \caption{Road debris models spawned by a simulator and the corresponding segmentation masks.}
    \label{fig:road_debris_models}
\end{figure}

\subsection{Domain randomization on objects}\label{sec:domain_randomization}
\begin{figure}[!t]
    \centering
    \begin{subfigure}[b]{0.49\linewidth}
    \includegraphics[width=\linewidth]{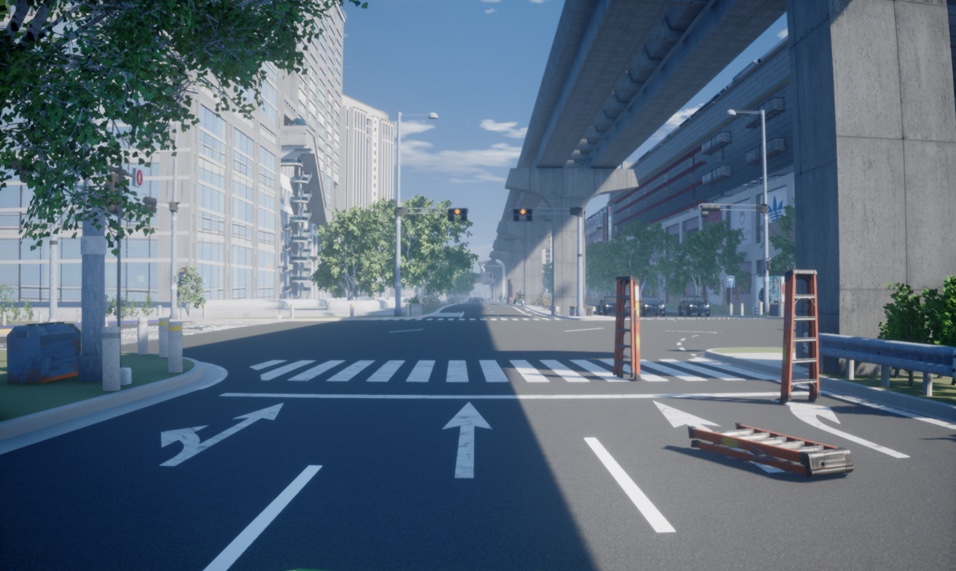}
    \end{subfigure}
    \begin{subfigure}[b]{0.49\linewidth}
    \includegraphics[width=\linewidth]{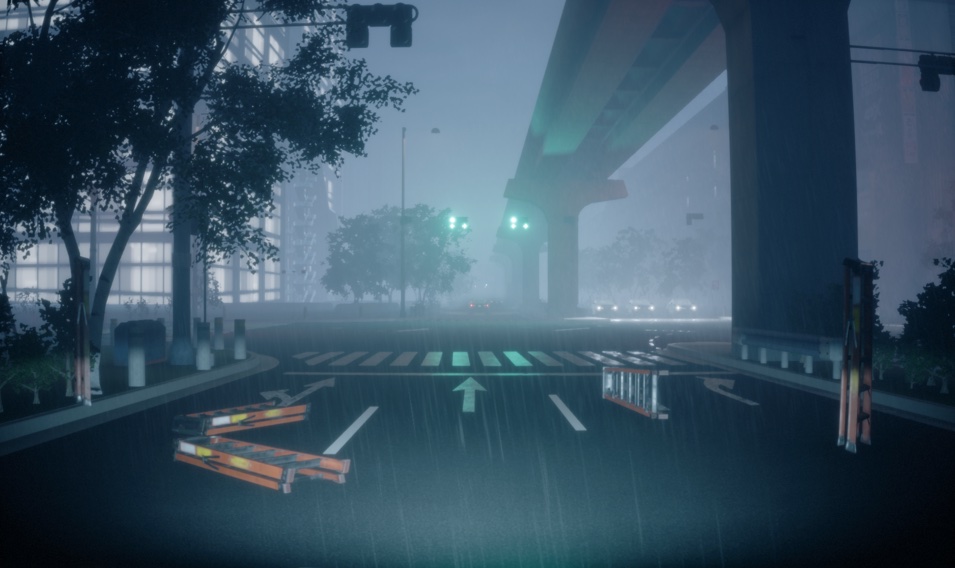}
    \end{subfigure}
    \hfill
    \caption{Examples of spawned ladder models using domain randomization with different times of day and weather conditions.}
    \label{fig:domain_randomization}
\end{figure}

The observable gap between synthetic models and real images includes appearances such as color, texture, and shadows. However, \cite{tremblay2018training} showed that for DNNs, such artifacts from synthetic images are less important than the actual shape of objects. Therefore, we apply domain randomization to the generated models in Section \ref{sec:3d_model_generation} by randomly positioning the models in the simulator with different color and textures.
Specifically, various road debris models in the simulator are generated by randomly sampling 3D position, 3D orientation, color tone, material, and visibility by fog or haze.
Once these models are generated, one or more of these domain randomized instances are spawned in a simulated environment with environmental conditions that match that of the real image being augmented (using the metadata described in Section \ref{sec:3d_model_generation}).
Examples of domain randomization are shown in Figure~\ref{fig:domain_randomization}.
Most of the aforementioned randomization hyper-parameters were selected in a uniformly random manner. 
For model position, objects were placed up to 300 meters away from the camera or when the height of the 2D bounding box is at least 10 pixels. 
For model orientation, yaw, pitch, and roll angles were randomly selected. This orientation was further automatically corrected to follow the physics rules in the simulator, which considers model shape, road shape, and gravity. 

\subsection{Semantic model augmentation}\label{sec:semantic}
\begin{figure}[ht]
    \centering
    \begin{subfigure}[b]{0.49\linewidth}
    \includegraphics[width=\linewidth]{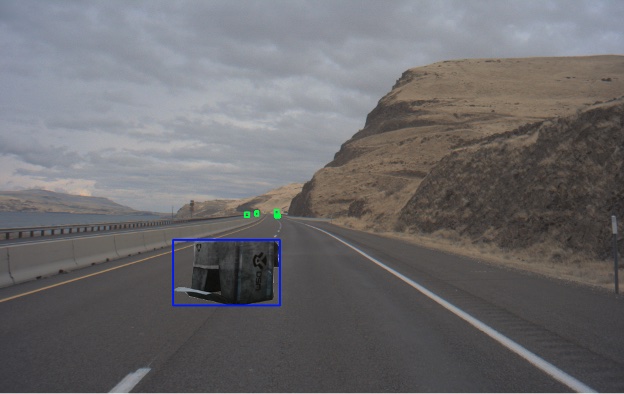}
    \end{subfigure}
    \begin{subfigure}[b]{0.49\linewidth}
    \includegraphics[width=\linewidth]{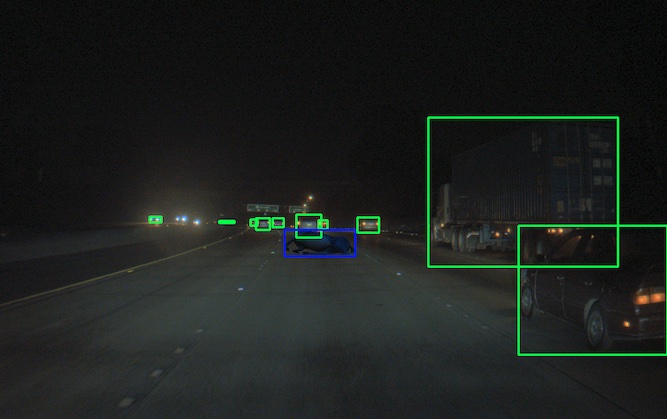}
    \end{subfigure}
    \hfill
    \begin{subfigure}[b]{0.49\linewidth}
    \includegraphics[width=\linewidth]{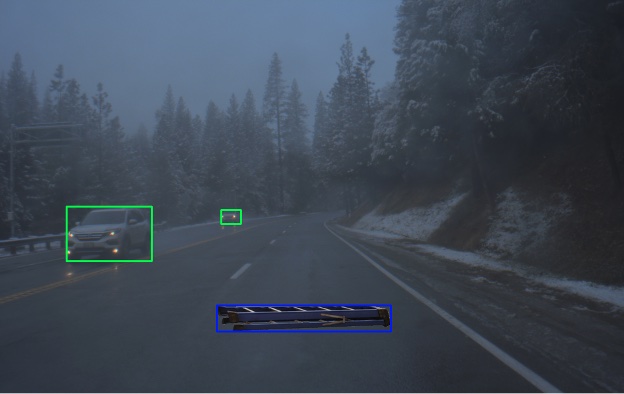}
    \end{subfigure}
    \begin{subfigure}[b]{0.49\linewidth}
    \includegraphics[width=\linewidth]{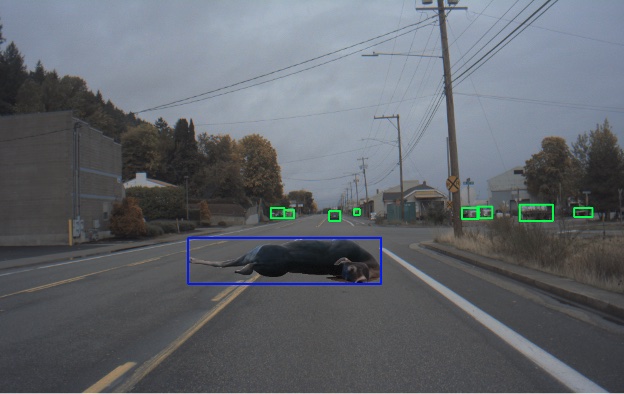}
    \end{subfigure}
    \hfill
    \begin{subfigure}[b]{0.492\linewidth}
    \includegraphics[width=\linewidth]{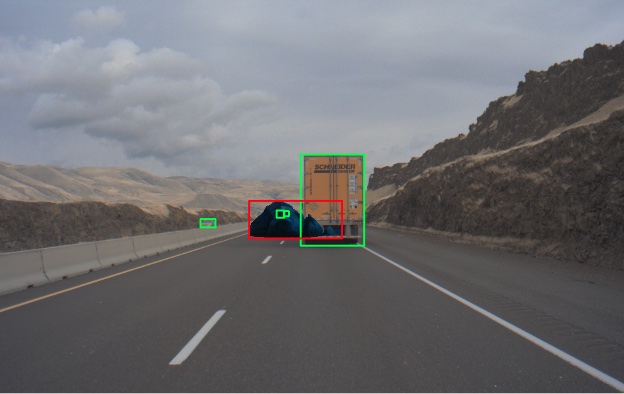}
    \end{subfigure}
    \begin{subfigure}[b]{0.49\linewidth}
    \includegraphics[width=\linewidth]{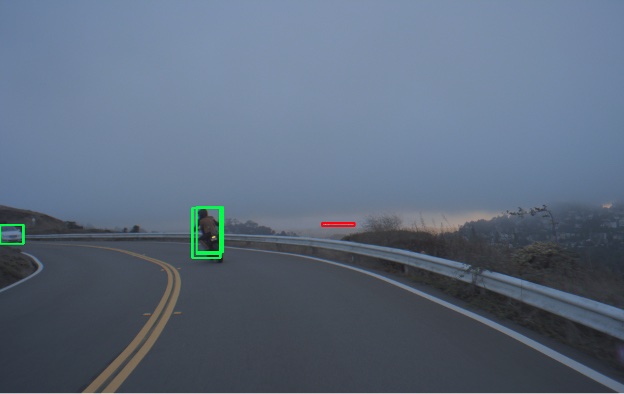}
    \end{subfigure}
    \hfill
    \vspace{-1.0em}
    \caption{Semantic model augmentation. Green rectangles: ground truth of dynamic objects labeled by a human. Blue rectangles: accepted semantic augmentation. Red rectangles: rejected augmentation.}
    \label{fig:semantic_augmentation_examples}
\end{figure}

In the last step of data generation, we augment the real images with the domain randomized synthetic models. During augmentation, we need to encode the semantic meaning of road debris, e.g.\ unusual objects on the road blocking a vehicle's path. By adding more semantic constraints for where road debris can be located, DNNs can more efficiently and accurately learn to distinguish road debris from other road elements.  Therefore, synthetic models are placed on the planned path of the ego vehicle or its neighboring lanes (either left and right lanes or shoulders). Models are not augmented in other unrelated lanes or in the sky to reduce the complexity of the road debris domain. Even though road debris can be on the sidewalk or in opposite lanes, debris out of the ego-vehicle's driving area are not considered for augmentation. In addition, we constrain the synthetic model to be on the ground and it cannot overlap with other objects identified by existing human-generated ground truth labels. 
Figure~\ref{fig:semantic_augmentation_examples} shows examples of semantic augmentation with various synthetic models and environmental conditions. The green rectangles are human-labeled ground truth, the blue rectangles indicate acceptable augmentations, and the red rectangles are rejected for augmentation. We note that synthetic models in accepted augmentations are on the road, and overlap with other objects is accepted as long as the synthetic model is in front of existing objects (e.g.\ any occlusion introduced by augmentation must be physically plausible). In the case of the rejected augmentation shown in the bottom left image in Figure~\ref{fig:semantic_augmentation_examples}, the synthetic rock model is unnaturally above the truck. This case was filtered out by checking the y-coordinates of the bottom of overlapping rectangles. In the bottom right image, the synthetic wood pallet model was not an accepted augmentation since it was placed in the sky. 

 \vspace{-0.4em}
 
\subsection{HazardNet architecture}\label{sec:hazardnet}
\begin{figure}[ht]
    \centering
    \includegraphics[width=70mm]{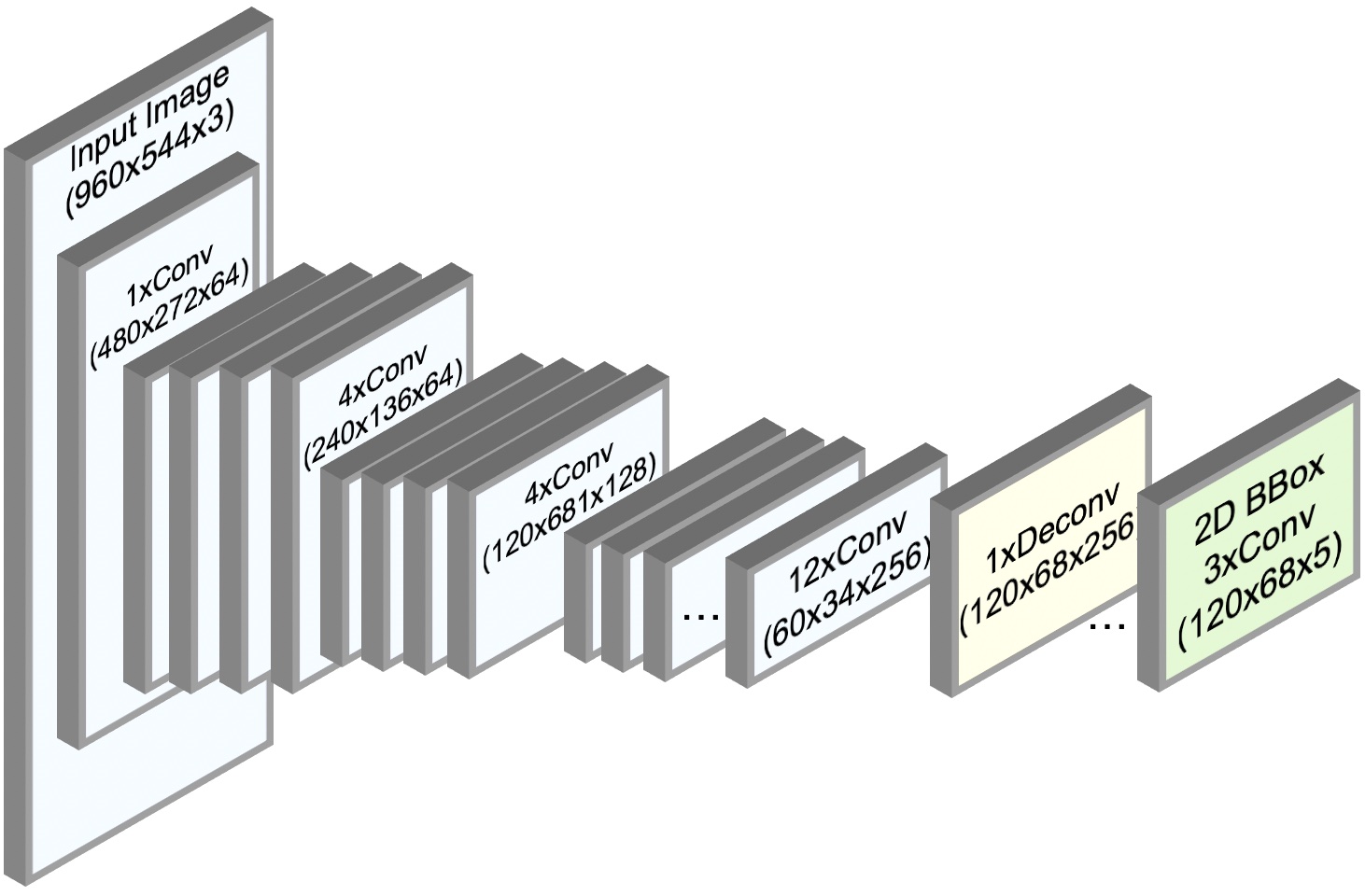}
    \caption{Architecture of HazardNet consisting of 263 layers and based on ResNet \cite{he2016deep}.}
    \label{fig:hazardnet_arch}
\end{figure}

ResNet \cite{he2016deep} is used as the backbone architecture of HazardNet. However, DarkNet \cite{redmon2017yolo9000} or any DNN can be used since the main contribution of this paper is balanced data collection and sampling, domain randomization, and semantic augmentation. HazardNet learns features and outputs bounding box proposals, which are post-processed using non-maximum suppression (NMS) to output the final detection results. The architecture of HazardNet is shown in Figure~\ref{fig:hazardnet_arch}. 
The final output layers ($120\times68\times5$) consist of one channel for detection confidence regression and four channels for bounding box positions, e.g.\ the normalized center position $(x_c, y_c)$ and the width $w$ and height $h$ both divided by two. 
Binary cross entropy (BCE) in Equation 
(\ref{eq_binary_cross_entropy}) is used as the loss function of the confidence channel for each output pixel. 
\begin{equation}\label{eq_binary_cross_entropy}
BCE(t, p) = t\cdot log(p)+(1-t) \cdot log(1-p-\epsilon)
\end{equation}
where $p$ is the predicted value of from the DNN,  $t$ is the corresponding ground truth value, and 
$\epsilon$ (we used $10^{-7}$) is to avoid numerical error when $p=1$. Since real data was sampled balancing over micro operational design domains ($\mu$ODD) as discussed in Section \ref{sec:real_data}, the focal loss function \cite{Lin_2017_ICCV} was not used.  
The loss function of the bounding box channels is the $l_1$ norm in Equation (\ref{eq_l1_loss}).
\begin{equation}\label{eq_l1_loss}
l_1(\pmb{T}, \pmb{P}) = ||\pmb{T} - \pmb{P}||_1
\end{equation}
where $\pmb{T} = [x_c, y_c, w/2, h/2]$ is the ground truth bounding box and $\pmb{P}$ is the prediction. 
The network is further optimized with an INT8 representation using TensorRT\footnote{\url{https://developer.nvidia.com/tensorrt}}. The inference time of an optimized HazardNet model, including post-processing times, is less than $5ms$ using an Nvidia RTX GPU. Therefore, more than $200$ frames of $(960\times544\times3)$ resolution images can be processed in a second. 

HazardNet is trained using the procedure in Figure~\ref{fig:workflow}. The developed pipeline is for production and all processes are automated without human intervention. Currently, synthetic models are semantically augmented into real images offline. However, we are working on completing a whole training pipeline for online semantic data augmentation in the training stage using a new real-time simulator. When training is complete, we evaluate the model using real road debris test data (see e.g.\ Section \ref{sec:test_datasets}). We then extract all false positive (FP) and false negative (FN) cases and create a new set of synthetic data imitating the failure cases. Since road debris in the real test data do not have matching synthetic models, we generate data based on the 3D size of road debris. 3D information is always available since all real-world objects are labeled using both cameras and lidar. We continue this cyclic workflow until the model performance (based on FP and FN) converges, as shown in Figure~\ref{fig:cyclic_flow}. 


\subsection{Performance metrics}\label{sec:metrics}
Detectors trained to perform few-shot learning tend to output lots of false positive detections \cite{zhu2020don}. 
However, in road debris detection it is crucial to minimize false positive detection. One of the most important objectives of the road debris detector is thus the false positive detection rate (FPR): 
$ FPR = FP / (FP + TN)$, where $FP$ is the number of false positive detections and $TN$ is the number of true negative detections. 
When road debris is detected, the vehicle needs to be stopped and sometimes may be required to brake suddenly. Harsh braking due to a false positive detection can cause severe accidents with neighboring vehicles on highways and thus must be avoided. Therefore, the FPR of HazardNet should be significantly low. Achieving both high recall and low FPR is challenging, so data augmentation using well-balanced, large sets of real data is key. Since the real data we use is collected from numerous operational design domains (ODD) and contains most road entities in a variety of scenarios, HazardNet is able to learn unusual road debris against common road entities. We examine the effects of training and evaluating on real data in the next section.


\section{Experiments}

\subsection{Data collection and labeling}\label{sec:real_data}
Among publicly available datasets, the nuScenes dataset \cite{caesar2020nuscenes} includes some images of traffic cones and barriers, which are categorized as hazards but common road elements with enough training data.  \cite{pinggera2016lost} provides open road debris data for stereo vision but we were not able to use it since their test data includes our synthetic models such as cardboard boxes, tires, logs, traffic cones, wood pallets. 
Therefore, we collected our own real road debris dataset for evaluation (i.e.\ the Real training and test datasets described in Sections \ref{sec:training_datasets} and \ref{sec:test_datasets}, respectively).
In addition, we used a well-balanced, in-house real dataset to apply semantic augmentation (i.e.\ for the Sim training and test datasets described in Sections \ref{sec:training_datasets} and \ref{sec:test_datasets}).

First, real data without load debris was collected for Sim data. It was collected from various locations, lighting conditions, times of day, and weather conditions. There are four main axes of data collection buckets: 
\begin{itemize}
    \item Road type: highways, freeways, suburban roads, urban roads, rural roads, and dirt roads, indoor/outdoor parking lots.
    \vspace{-0.5em}
    \item Time of day: day, night, dawn/dusk, sunset/sunrise
    \vspace{-0.5em}
    \item Weather: clear, sun, moon, cloud, rain, snow, fog
    \vspace{-0.5em}
    \item Objects: passenger car, emergency vehicles, heavy trucks, bicycle, motor bike, scooter, pedestrians with different level of traffic 
\end{itemize}
Even data was collected considering above guidelines, the majority of data tends to be  in clear day time on a straight road. When data is not balanced for each category, under-represented categories tend to fail. For example, when uncommon construction trucks or snowing weather are quite rare, the DNN model trained on biased data fails on detecting such as trucks. With such difficulties, data should be collected considering the balance of data. 

Secondly, the Real road debris were staged on the road in various locations and orientations. Staging was done because collecting real road debris data in-the-wild is extremely difficult, even with access to millions of customers' vehicle data. Collecting road debris data on highways is also dangerous and does not enable staging, though such settings are important to consider. As a result, the data was collected in a limited number of relatively static environments.
We placed 30 different road debris on private roads and a vehicle equipped with all sensors (camera, lidar, radar) recorded data starting from about 200 meters away and drove towards the road debris. The staged real road debris are all unseen categories and unseen objects excluded from synthetic models. The unseen road debris includes trash bags, stuffed animals, wood branches, standing barriers, folded cardboard boxes, delineator posts, and more. 



For all the real data used in this paper, the ego-vehicle was equipped with not only cameras but also lidar and radar sensors. When human labelers annotate objects and assign corresponding 3D distances from the vehicle, these radar and lidar sensors are fully utilized for more accurate 3D estimation of roads and objects. For every camera frame, human labelers annotated dynamic objects such as vehicles, bikes, and pedestrians, as well as static objects such as lane lines, road marks, road boundaries, traffic signs/lights, and vertical landmarks using all sensors. 

Every image was also associated with other metadata such as weather, time/date, refined GPS/IMU signals (latitude, longitude, altitude, and orientation), and road condition (wet, snow, dirt, etc.). As mentioned in Section \ref{sec:3d_model_generation}, this information is used when augmenting real images with synthetic models. 

\subsection{Training datasets}\label{sec:training_datasets}
\begin{figure}[ht]
    \vspace{-1em}
    \centering
    \includegraphics[width=60mm]{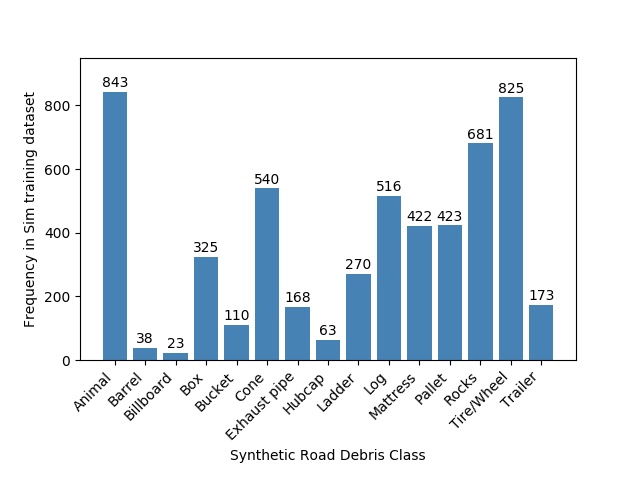}
    \caption{Frequency of synthetic model classes in the Sim training dataset.}
    \vspace{-1.5em}
    \label{fig:sim_data_frequency}
\end{figure}

The experiments were conducted using three different training datasets, which we refer to as: Sim, Real, and Hybrid(Sim + Real). The Sim training dataset consists of about 5,000 sampled images balanced on various $\mu$ODD as described in Section \ref{sec:real_data}. The 3D synthetic road debris models were used to augment the images with domain randomization and semantic augmentation. The breakdown of each synthetic road debris subclass is shown in Figure \ref{fig:sim_data_frequency}. The Real training dataset consists of about 10,000 real images and has no overlap with the Sim dataset images. The road debris in each Real dataset image was labeled by humans as described in Section \ref{sec:real_data}. The Hybrid training dataset is the combination of the Sim and Real training datasets, and thus has a total of about 15,000 images.


\subsection{Test datasets}\label{sec:test_datasets}
We evaluated HazardNet on two test datasets (not used in training): Real and Sim, which were used for cross-validation of the model trained using few-shot learning and the model with fully supervised learning, respectively.
The Real test dataset consists of about 3,000 images, and each image contains zero, one or multiple labeled road debris instances with various poses (see Section \ref{sec:real_data} for details).
This real, unseen test dataset is used to evaluate the performance of HazardNet for few-shot learning.
The Sim test dataset is generated in the same manner as the training dataset using domain adaptation and semantic augmentation of synthetic models on real images. The test data consists of about 5,000 images. This Sim test data is used to evaluate supervised learning of HazardNet and represents an upper bound on performance.  

\begin{table*}[ht]
    \centering
    \begin{tabular}{|*{9}{c|}}
    \hline
    \textbf{mAP(\%)} & \multicolumn{4}{c}{\textbf{Sim Test Dataset}} & \multicolumn{4}{|c|}{\textbf{Real Test Dataset}} \\\hline
    \textbf{Train Dataset} & Small & Medium & Large & All & Small & Medium & Large & All\\
    \hline
    Sim$^a$ & 96.05 & 99.56 & 100 & 97.79 & 35.11 & 41.44 & 6.16 & \textbf{35.42}$^b$\label{tab:sim2real} \\
    \hline
    Real$^c$ & 6.89 & 9.73 & 0.10 & 4.22 & 73.92 & 83.76 & 75.27 & 76.36$^d$ \\
    \hline
    Hybrid(Sim + Real) & 95.03 & 99.13 & 100 & 97.13 & 74.79 & 91.78 & 80.41 & 80.53 \\
    \hline
    \end{tabular}
    \\
    \footnotesize{$^a$ Sim is real image data with synthetic model augmentation, $^b$ mAP of few-shot learned HazardNet,\\ $^c$ Real is staged real road debris data,  $^d$ mAP of the supervised learned  network}
    \caption{Mean average precision(mAP) scores using 0.5 IOU for HazardNet trained on Sim, Real, and Sim + Real training datasets. The models were evaluated on both the Sim and Real test datasets and the mAP scores are further divided into small, medium, and large road debris.}
    \label{tab:train_experiments}
\end{table*}

\subsection{Quantitative evaluation}

To quantify the performance of HazardNet trained on the three training datasets, Sim, Real, and Hybrid (Sim+Real) described in Section \ref{sec:training_datasets}, we compute the mean Average Precision(mAP), true positive rate (TPR), false positive rate (FPR), precision, and recall. For mAP, we divide each of the two test datasets into difficulty buckets: small (8-25 pixel height), medium (25-100 pixel height), large (100+ pixel height), and all. When evaluating all the test instances, we weight the mAP corresponding to each bucket proportional to object size. The corresponding weights for small, medium, and large objects are $0.5$, $1$, and $5$.
Table \ref{tab:train_experiments} shows cross-validation results for the Sim and Real training and tests: (1) sim2sim($97.79\%$) and real2real($76.36\%$) are the results of supervised learning with Sim and Real data, respectively. As expected, these supervised learning approaches result in high performance. (2) sim2real (trained on Sim and tested on Real data) is the proposed few-shot learning framework using HazardNet with $\pmb{35.42\%}$ mAP in the All category. The low performance in the large category (6.16\%) is due to the limited number of augmented images with objects close to the ego-vehicle, since we mostly focus on detecting distant road debris. (3) The other cross-validation, real2sim (4.22\%), provided the worst results. Since the Real dataset is staged in limited venues (as described in Section \ref{sec:real_data}), the network overfit to those environments and was not able to extrapolate to the Sim dataset domain. (4) hybrid2real is a more interesting case. We combined Sim and Real data and tested on Real data, and the Hybrid model (hybrid2real, 80.53\%) outperformed the supervised learning approach (real2real, 76.36\%). These results imply that the few-shot learned sim2real and hybrid models are more generative than the real2real model with supervised learning. 

\begin{figure}[ht]
    \centering
    \includegraphics[width=65mm]{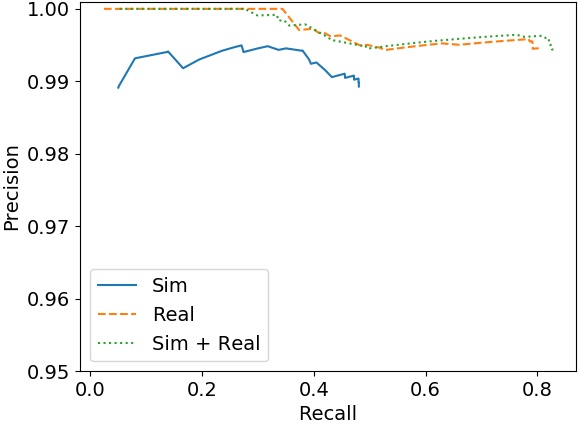}
    \caption{Precision-Recall curves for models evaluated on the Real test dataset.}
    \vspace{-0.8em}
    \label{fig:real_prec_recall}
\end{figure}

\begin{figure}[ht]
    \centering
    \includegraphics[width=65mm]{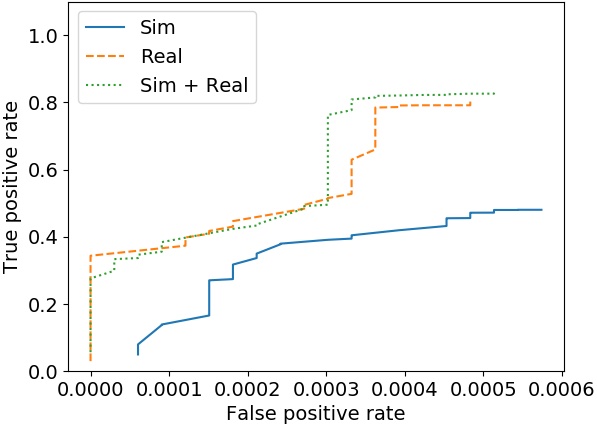}
    \caption{ROC curves for models evaluated on the Real test dataset.}
    \label{fig:real_roc}
\end{figure}

Figure \ref{fig:real_prec_recall} plots precision-recall curves for all three models evaluated on the Real test dataset. All three cases have substantially high precision (more than 99\%) even though the few-shot learned Sim model had lower recall. The confidence threshold value was determined on the validation dataset with the highest F-score, which was 0.3 for the Sim model. In addition, Figure \ref{fig:real_roc} plots the receiver operating characteristic (ROC) curves corresponding to all three models evaluated on the Real test dataset. We observe that all three models have very low false positive detection rates (less than 0.06\%), which is essential for road debris detection even though the few-shot learned Sim model was not as good as the Sim or Hybrid models.  

\vspace{-1.5em}
\vspace{-1.5em}



\begin{figure*}[!htb]
\centering
\includegraphics[width=0.238\linewidth]{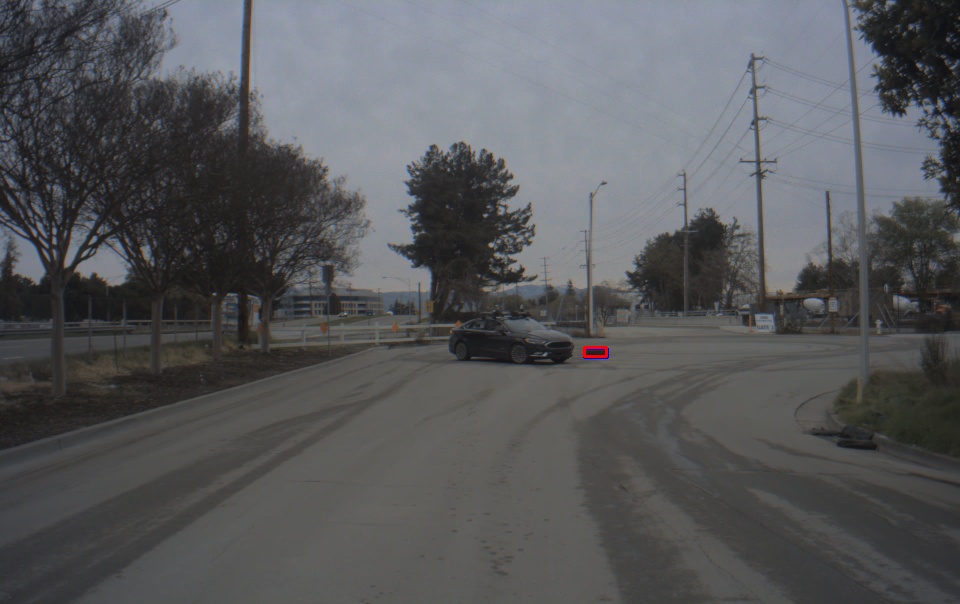}
\includegraphics[width=0.238\linewidth]{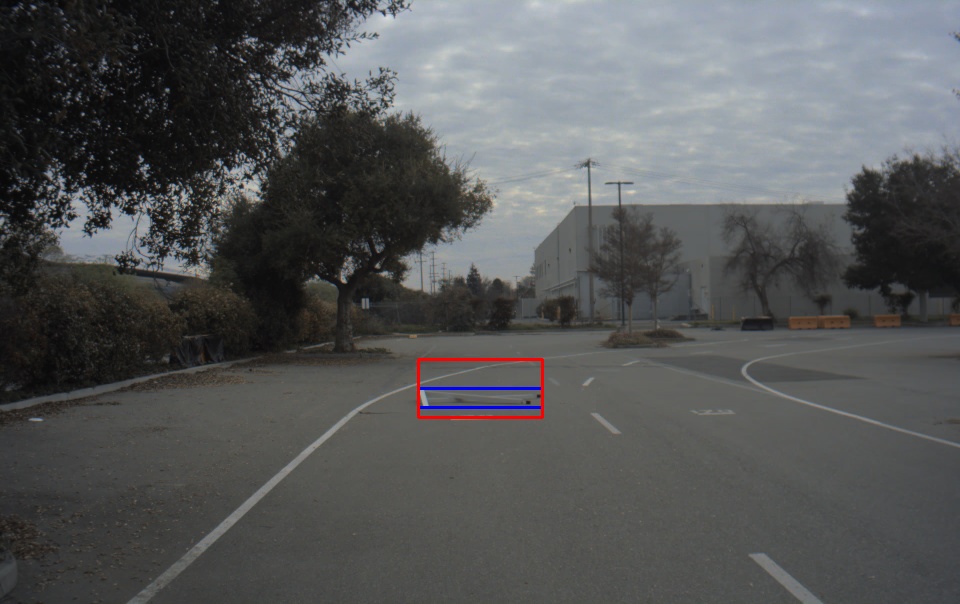}
\includegraphics[width=0.238\linewidth]{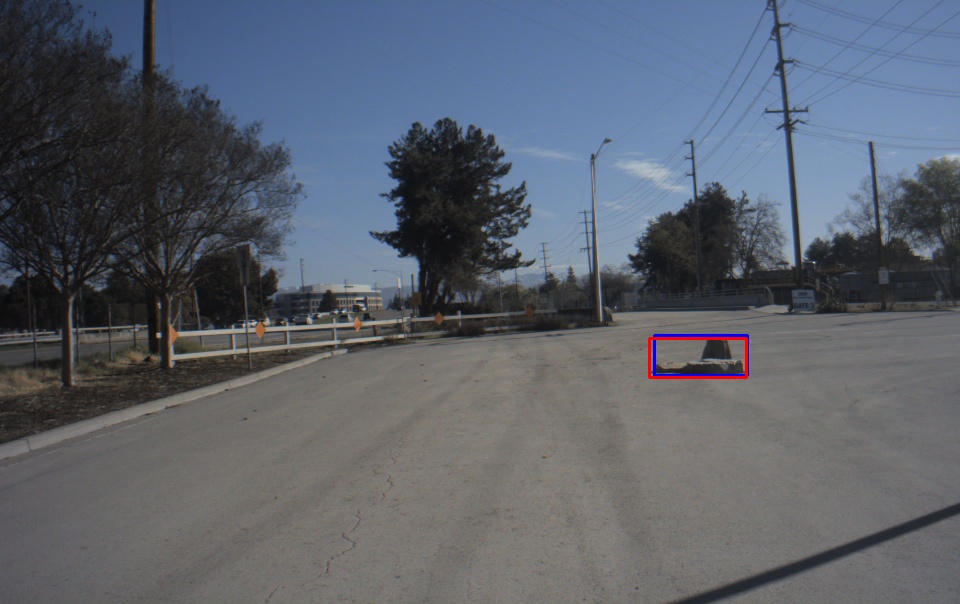}
\includegraphics[width=0.238\linewidth]{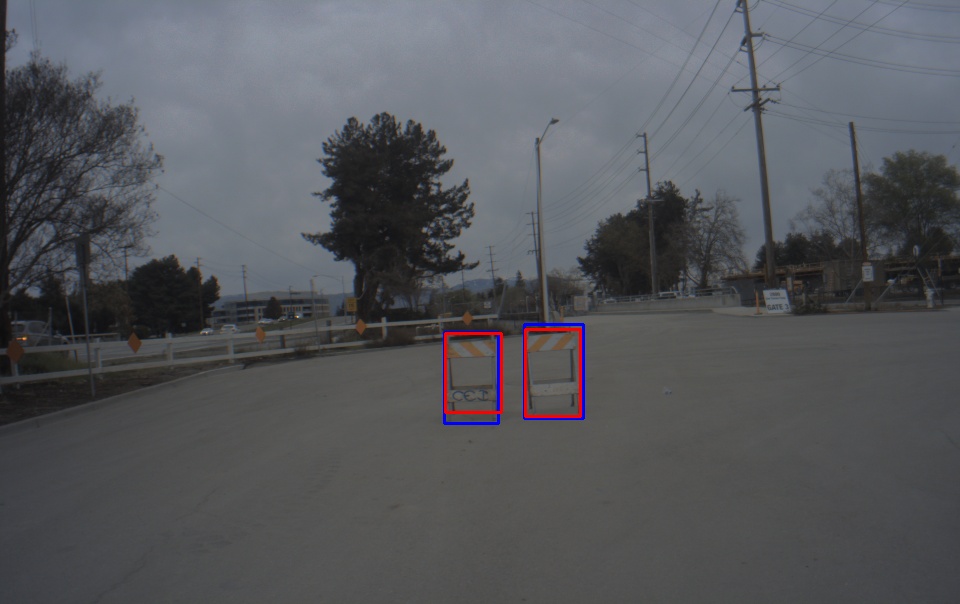}
\break
\includegraphics[width=0.238\linewidth]{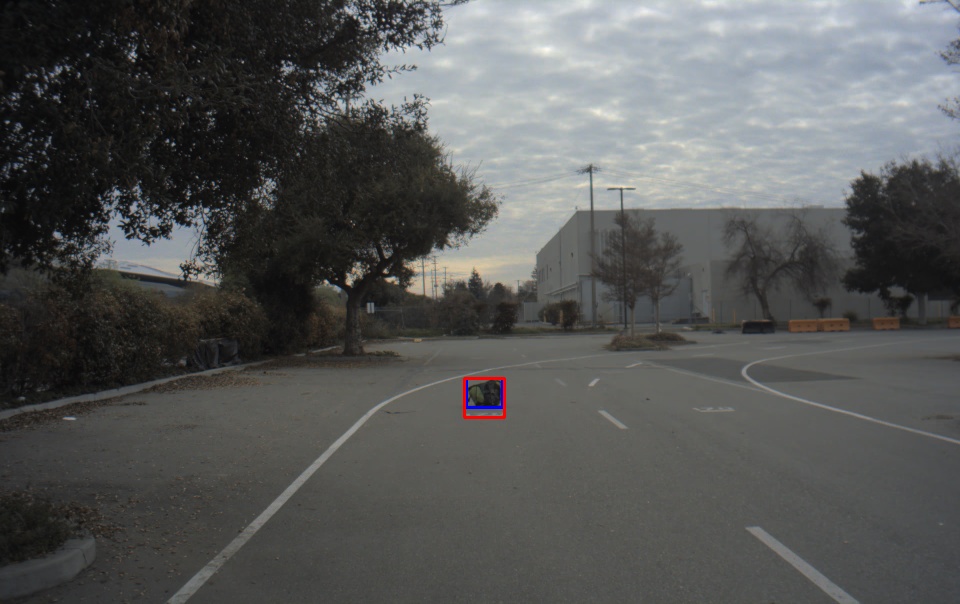}
\includegraphics[width=0.238\linewidth]{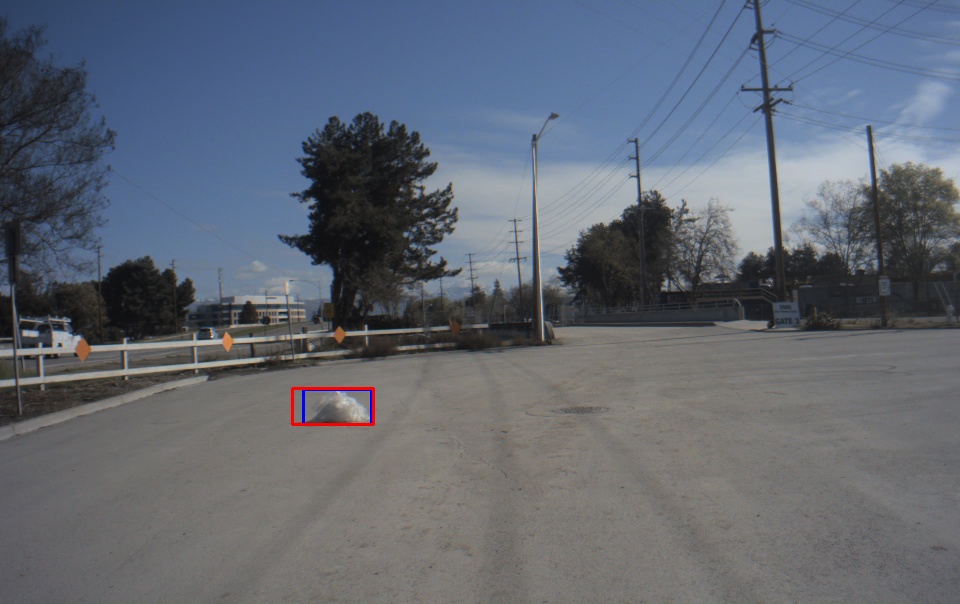}
\includegraphics[width=0.238\linewidth]{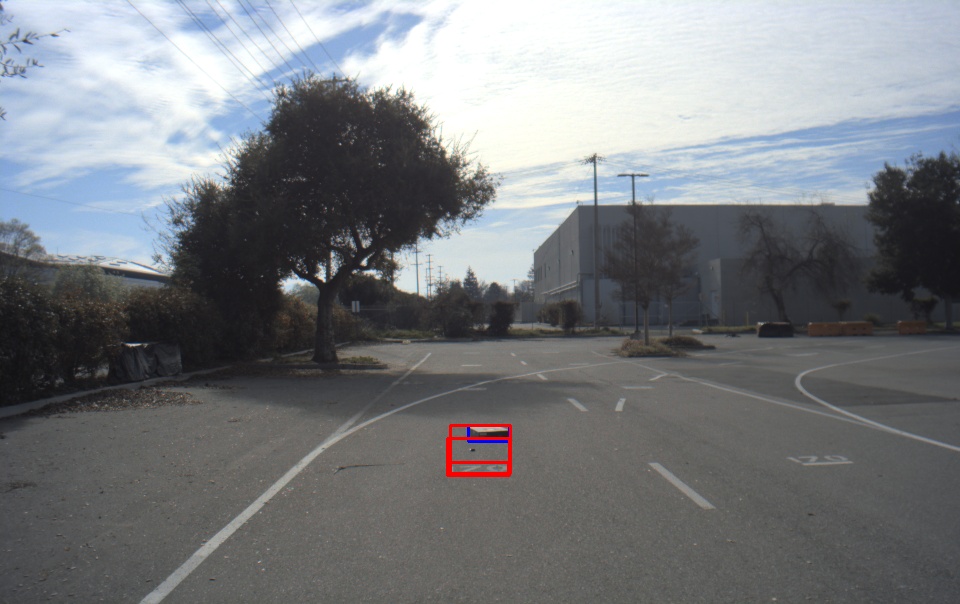}
\includegraphics[width=0.238\linewidth]{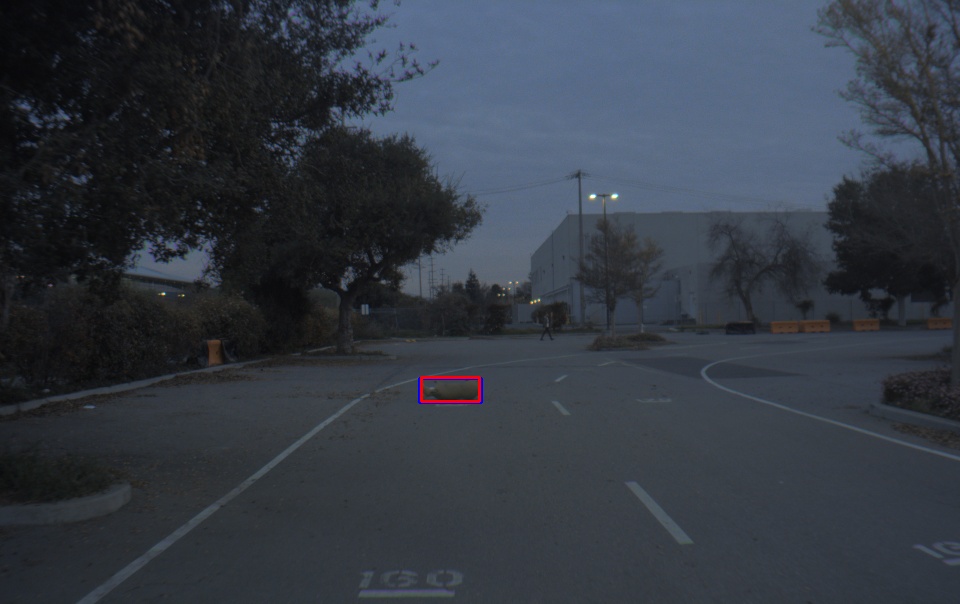}
\break
\includegraphics[width=0.238\linewidth]{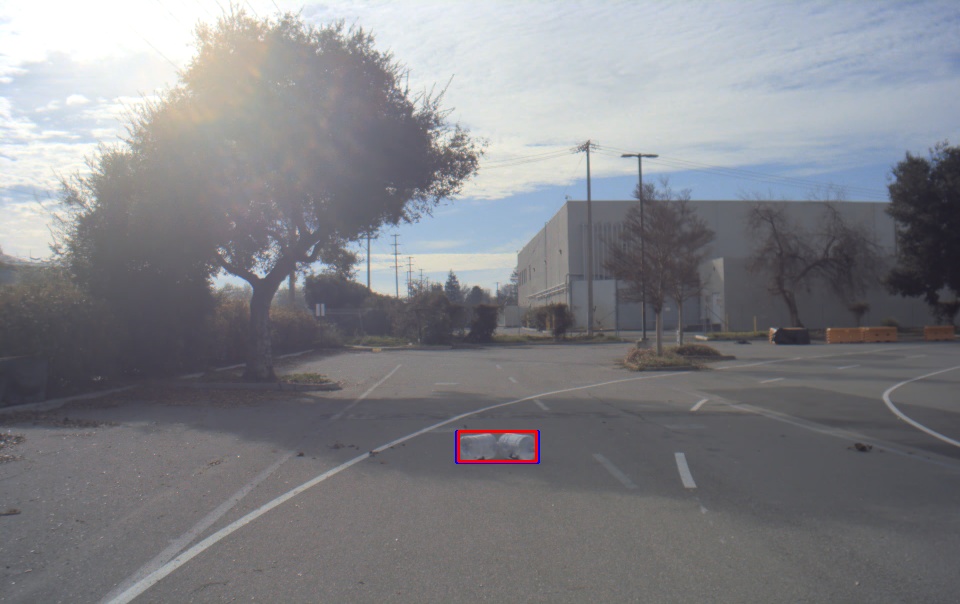}
\includegraphics[width=0.238\linewidth]{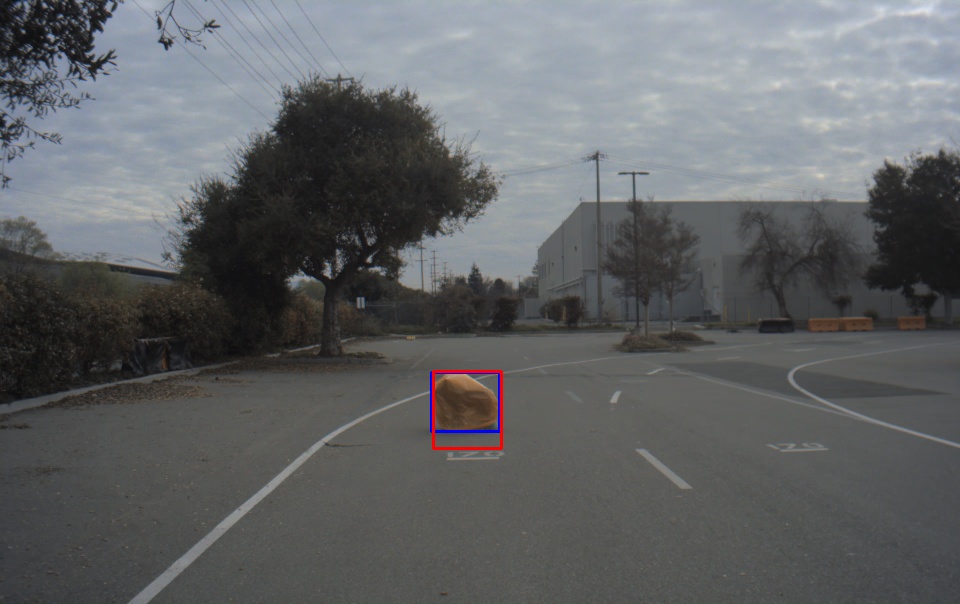}
\includegraphics[width=0.238\linewidth]{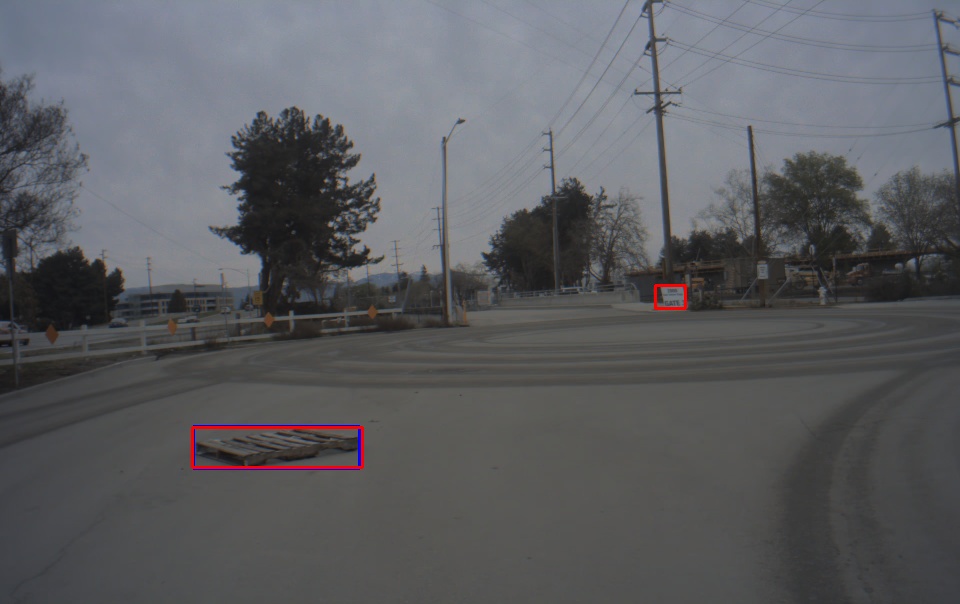}
\includegraphics[width=0.238\linewidth]{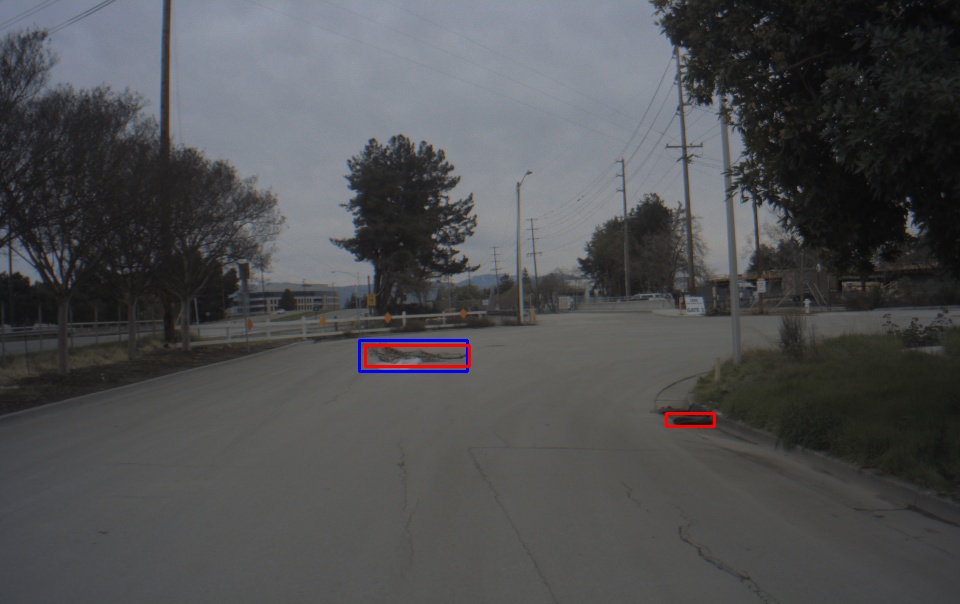}
\break
\caption{Detections of unseen road debris using HazardNet. Ground truth labels are drawn in blue, and HazardNet detections are drawn in red. The network is able to detect debris missed by human annotators (row 3, columns 3 and 4).}
\label{fig:detection_results}
\end{figure*}

\begin{figure*}[!htb]
\centering

\includegraphics[width=0.238\linewidth]{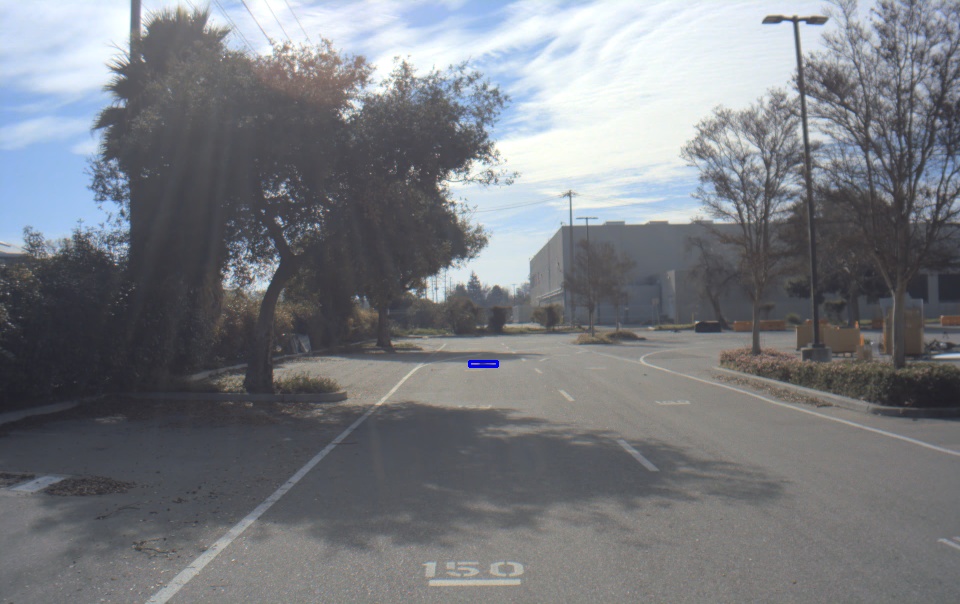}
\includegraphics[width=0.238\linewidth]{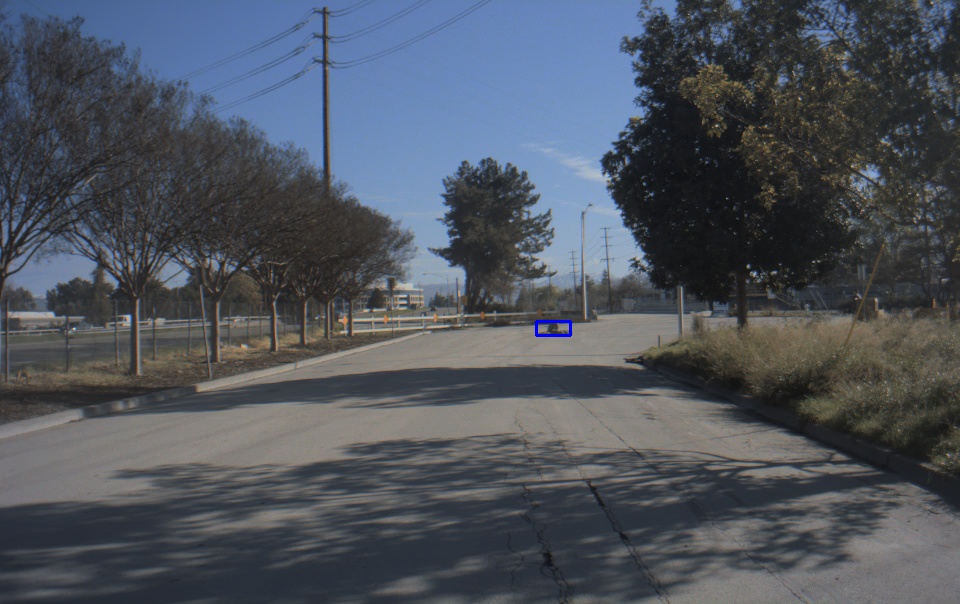}
\includegraphics[width=0.238\linewidth]{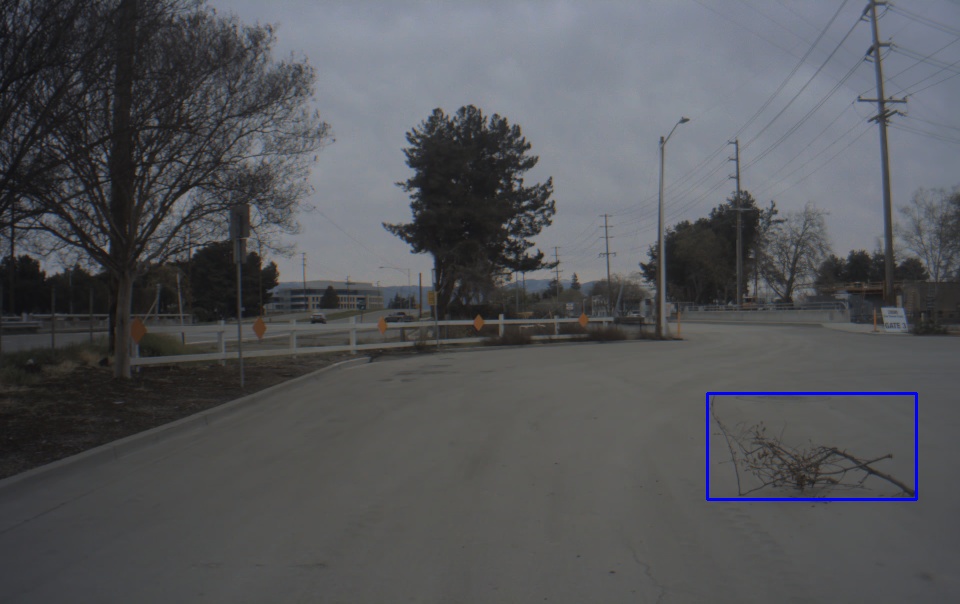}
\includegraphics[width=0.238\linewidth]{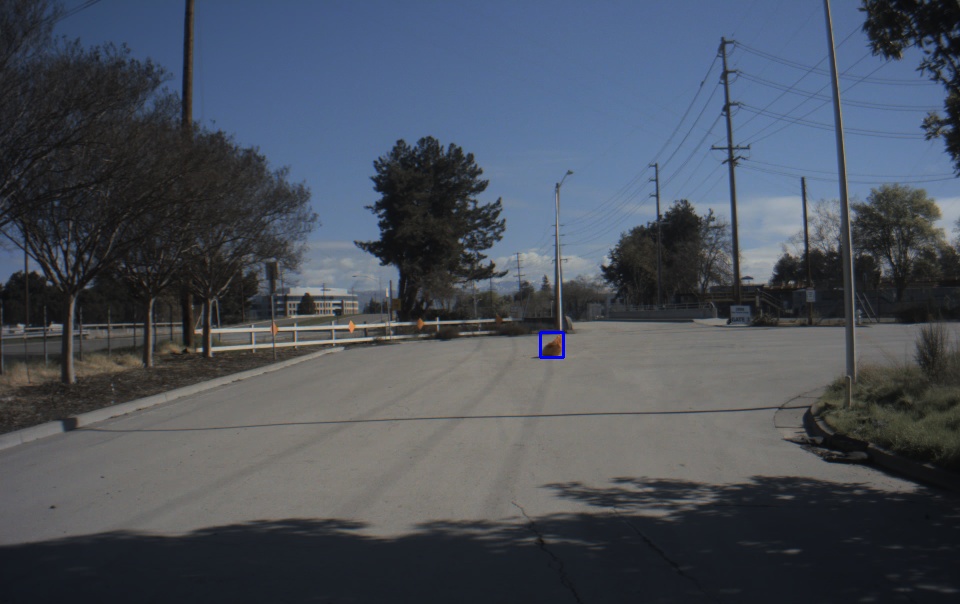}
\break
\caption{False negative detections of unseen road debris using HazardNet. Ground truth labels are drawn in blue, and HazardNet detections are drawn in red. Most of the false negative detections correspond to distant objects.}
\label{fig:fn_detection_results}
\end{figure*}

\vspace{3em}
\subsection{Qualitative evaluation}
To evaluate the efficacy of HazardNet in achieving few-shot learning, we also visualize detection results for real images containing unseen road debris (sim2real case). Figure \ref{fig:detection_results} shows example detections for a variety of road debris classes and environmental conditions. While HazardNet was trained only on simulated road debris, the network is able to detect real debris from both small and large distances from the camera. In Figure \ref{fig:fn_detection_results}, we also show examples of false negative detections.

\section{Conclusion}

We proposed a novel few-shot learning framework and deep learning model, HazardNet, which detects road debris for autonomous driving applications. We show that a small set of synthetic models can guide the DNN to learn to detect unseen real-world road debris.
Our method can be extended to other applications where large-scale, real datasets are hard to acquire.
During augmentation, the shadow of objects and the direction of lights were not considered and they were left for future research.  

{\small
\bibliographystyle{ieee_fullname}

}
\end{document}